\documentclass{article} 
\usepackage{nips11submit_e,times}

\usepackage{amsmath}
\usepackage{amsthm}
\usepackage{bm}
\usepackage{dsfont}
\usepackage{rotating}
\usepackage{multirow}
\usepackage{rotating}

\usepackage{amsmath}
\usepackage{amssymb}
\usepackage{bm}
\usepackage{amsthm}
\usepackage{graphicx}
\usepackage{dsfont}
\usepackage{algorithm}
\usepackage{algorithmic}
\usepackage{lscape}
\usepackage{rotating}
\usepackage{color}
\usepackage{url}

\title{Convergent Expectation Propagation in Linear Models with Spike-and-slab Priors}

\author{
Jos\'e Miguel Hern\'andez-Lobato\\
Department of Engineering\\
University of Cambridge\\
Trumpington Street, Cambridge \\
CB2 1PZ, United Kingdom\\
\texttt{jmh233@eng.cam.ac.uk} \\
\And
Daniel Hern\'andez-Lobato\\
Computer Science Department\\
Universidad Aut\'onoma de Madrid\\
Francisco Tom\'as y Valiente, 11\\
28049 Madrid, Spain\\
\texttt{daniel.hernandez@uam.es}
}

\nipsfinalcopy 

\begin{document}

\maketitle

\begin{abstract}
Exact inference in the linear regression model with spike and slab priors is 
often intractable. Expectation propagation (EP) can be used for approximate 
inference. However, the regular sequential form of EP (R-EP) may fail
to converge in this model when the size of the training set is very small.
As an alternative, we propose a provably convergent EP algorithm (PC-EP).
PC-EP is proved to minimize an energy function which, under some constraints, is bounded from below and
whose stationary points coincide with the solution of R-EP. 
Experiments with synthetic data indicate that when R-EP does not
converge, the approximation generated by PC-EP is often better. 
By contrast, when R-EP converges, both methods perform similarly.
\end{abstract}

\section{Introduction}

Exact Bayesian inference is often intractable in many probabilistic models of practical interest.
The computational cost of marginalization operations scales exponentially in the number of variables
or these operations require to compute high-dimensional integrals that do not have a closed-form analytical solution.
In practice, we have to use some form of approximation.
Approximate inference in large applications is frequently implemented using deterministic methods.
These have often less computational cost than other alternatives such as sampling.
Deterministic methods approximate the exact posterior by a tractable parametric distribution whose parameters
are selected by solving optimization problems.
Expectation propagation (EP) is one of the most successful techniques for deterministic
approximate inference \cite{Minka2001}. However, a disadvantage of EP is that, in its standard sequential form,
it is not guaranteed to converge.

In this work we focus on the linear regression model with spike-and-slab priors (LRMSSP). 
The standard sequential EP method often generates in this model very accurate approximations of the posterior distribution \cite{jmhlPhd2010}.
However, EP may fail to converge in some extreme cases in which the number of training instances is very small.
To avoid this, we introduce a provably convergent EP algorithm (PC-EP) for approximate inference in the LRMSSP.
PC-EP is based on the double loop algorithm described in \cite{IMM2005-03460}.
Each outer iteration of PC-EP is proved to minimize an energy function whose stationary points coincide with
the solution of regular EP \cite{Minka01}. The main difference between PC-EP and the algorithm proposed in
\cite{IMM2005-03460} is that PC-EP constrains the variance parameters to be positive.
This ensures that the energy function minimized by PC-EP is bounded from below.
The boundedness of this function is a necessary condition to guarantee convergence.
Experiments with synthetic data illustrate the advantages of PC-EP over regular EP. 
In these experiments, when regular EP does not converge, the posterior approximation generated by PC-EP is often better.
By contrast, when regular EP converges, both methods perform similarly.

\section{The Linear Regression Model with Spike-and-slab Priors}

We focus on the problem of Bayesian inference 
in the linear regression model with spike-and-slab priors (LRMSSP). 
Consider $n$ feature vectors, with $d$ real components each, encoded by the design matrix $\mathbf{X}=(\mathbf{x}_1,\ldots,\mathbf{x}_n)^\text{T}$
and associated target values $\mathbf{y}=(y_1,\ldots,y_n)^\text{T}\in \mathds{R}^n$. The LRMSSP
assumes that $\mathbf{y}  = \mathbf{X}\mathbf{w} + \bm \epsilon$,
where $\mathbf{w}$ is an unknown $d$-dimensional vector of real coefficients and $\bm \epsilon$ is
an $n$-dimensional vector of independent Gaussian noise with variance $\sigma^2$.
Given $\mathbf{X}$ and $\mathbf{y}$, the likelihood of $\mathbf{w}$ is a Gaussian function:
$\mathcal{P}(\mathbf{y}|\mathbf{w},\mathbf{X}) = \mathcal{N}(\mathbf{y}|\mathbf{X}\mathbf{w},\sigma^2 \mathbf{I})$.
The prior for $\mathbf{w}$ is a product of spike-and-slab factors \cite{0884.62031}:
\begin{align}
\mathcal{P}(\mathbf{w}) & = \prod_{i=1}^d \left[ p_s \mathcal{N}(w_i|0,v_s) + (1-p_s) \delta(w_i) \right]\,,
\label{eq:spike_and_slab}
\end{align}
where $\mathcal{N}(\cdot|0,v_s)$ denotes a Gaussian density with zero mean and variance $v_s$ (the slab), $\delta(\cdot)$ is
a delta function centered at zero (the spike), and $p_s$ is the prior probability that any of the components of $\mathbf{w}$ is different from zero. 
The LRMSSP has special practical interest in regression problems with more features than training instances (that is, $d \gg n$).
In this scenario, the assumption of sparsity is used to reduce over-fitting by limiting the number of influential features \cite{johnstone2009}. 
The spike-and-slab prior (\ref{eq:spike_and_slab}) incorporates this assumption by inducing a bi-separation in the model coefficients.
A small number of coefficients are considered to be different from zero. These are the coefficients sampled from the slab.
At the same time, a reduced number of coefficients are considered to be exactly zero. These are the coefficients sampled from the spike.

Given $\mathbf{X}$ and $\mathbf{y}$, the posterior distribution for $\mathbf{w}$ can be computed using Bayes' rule:
\begin{equation}
\mathcal{P}(\mathbf{w}|\mathbf{X},\mathbf{y}) = \frac{\mathcal{P}(\mathbf{y}|\mathbf{w},\mathbf{X}) \mathcal{P}(\mathbf{w})}{\mathcal{P}(\mathbf{y}|\mathbf{X})}\,.
\label{eq:Bayes}
\end{equation}
The central operation in the application of Bayesian methods is the computation of marginalizations or expectations with respect to 
this distribution. However, these operations are usually intractable. As a solution,
we use EP \cite{Minka01} to produce a deterministic approximation to (\ref{eq:Bayes}) which is tractable.
The numerator in the right part of (\ref{eq:Bayes}) can be written as
\begin{align}
\mathcal{P}(\mathbf{y}|\mathbf{w},\mathbf{X}) \mathcal{P}(\mathbf{w}) = 
\mathcal{N}(\mathbf{y}|\mathbf{X}\mathbf{w},\sigma^2 \mathbf{I}) \prod_{i=1}^d f_i(w_i)\,,
\label{eq:unPosterior}
\end{align}
where $f_i(w_i) = p \mathcal{N}(w_i|0,v_s) + (1-p) \delta(w_i)$.
EP replaces each $f_i$ by a 
Gaussian factor $\tilde{f}_i(w_i) = \exp \left\{ \tilde{v}_{1i} w_i - \frac{1}{2} \tilde{v}_{2i} w_i^2 + \tilde{z}_i \right\}$,
where $\tilde{v}_{1i}$, $\tilde{v}_{2i}$ and $\tilde{z}_i$ are free parameters. Let $\mathcal{Q}(\mathbf{w})$ be the
product of the likelihood $\mathcal{N}(\mathbf{y}|\mathbf{X}\mathbf{w},\sigma^2 \mathbf{I})$ and the Gaussian factors $\tilde{f}_1,\ldots,\tilde{f}_d$.
Then $\mathcal{Q}$ is an unnormalized multivariate Gaussian distribution:
\begin{align}
\mathcal{Q}(\mathbf{w}) &= \mathcal{N}(\mathbf{y}|\mathbf{X}\mathbf{w},\sigma^2 \mathbf{I}) \prod_{i=1}^d 
	\tilde{f}_i(w_i) 
	\propto \mathcal{N}(\mathbf{w}|\mathbf{m},\mathbf{A}^{-1})\,, \label{eq:Q}
\end{align}
where $\mathbf{A}=\sigma^{-2} \mathbf{X}^\text{T} \mathbf{X} + \text{diag}(\tilde{\mathbf{v}}_2)$,
$\mathbf{m} = \mathbf{A}^{-1} (\tilde{\mathbf{v}}_1 + \sigma^{-2} \mathbf{X}^\text{T}\mathbf{y})$, $\tilde{\mathbf{v}}_1=(v_{11},\ldots,v_{1d})^\text{T}$
and $\tilde{\mathbf{v}}_2=(v_{21},\ldots,v_{2d})^\text{T}$.
Note that the posterior (\ref{eq:Bayes}) is obtained when (\ref{eq:unPosterior}) is normalized so that it integrates one.
Similarly, the EP approximation to the posterior is given by the normalized version of $\mathcal{Q}$.

The sequential implementation of EP iteratively updates the $\tilde{f}_i$ until reaching a stationary point
of a specific energy function \cite{Minka01}. However, the EP update operations do not guarantee the minimization
of this energy function and the method may sometimes fail to converge. 
Non-convergence can be prevented by \emph{damping} the EP update operations \cite{minka2002expectation}. 
After performing one update operation,
damping sets the updated factor $\tilde{f}_i$ to a log-convex combination of the factor before (old) and after (new) the update,
that is, $\tilde{f}_i=(\tilde{f}_i^\text{new})^\tau(\tilde{f}_i^\text{old})^{1-\tau}$, 
with $\tau \in [0,1]$. However, even when damping is used, EP can fail to 
converge in some extreme cases in which the number of training instances is very small.
When this happens, we can limit the maximum number of iterations of the algorithm 
and hope that the resulting approximation is accurate enough.
As an alternative, we present a provably convergent EP algorithm (PC-EP).

\section{Provably Convergent EP for the LRMSSP}

The EP solutions in the LRMSSP are in a one-to-one correspondence with stationary points of the following objective \cite{720257,Seeger2011}:
\begin{align}
	\underset{
	\begin{array}{c}
	v
	\end{array}
	}{\text{min}} 
	\underset{
	\begin{array}{c}
	\hat{v},\tilde{v}
	\end{array}
	}{\text{max}} 
E(v,\hat{v},\tilde{v})\,,\label{eq:objective}
\end{align}
where $v=\{\mathbf{v}_1,\mathbf{v}_2\}$, $\tilde{v}=\{\tilde{\mathbf{v}}_1,\tilde{\mathbf{v}}_2\}$,
$\hat{v}=\{\hat{\mathbf{v}}_1,\hat{\mathbf{v}}_2\}$, the elements in
these tuples are $d$-dimensional real vectors satisfying
\begin{align}
\mathbf{v}_1 & = \tilde{\mathbf{v}}_1 + \hat{\mathbf{v}}_1\,, &
\mathbf{v}_2 & = \tilde{\mathbf{v}}_2 + \hat{\mathbf{v}}_2\,,\label{eq:equality}
\end{align}
$E(v,\hat{v},\tilde{v})$ is an energy function given by
\begin{equation}
E(v,\hat{v},\tilde{v}) = - \log Z(\tilde{v}) - \log \hat{Z}(\hat{v}) + \log \tilde{Z}(v)
\end{equation}
and $Z(\tilde{v})$, $\hat{Z}(\hat{v})$ and $\tilde{Z}(v)$ are the following normalization constants:
\begin{align}
Z(\tilde{v}) &= \int \mathcal{N}(\mathbf{y}|\mathbf{X}\mathbf{w},\sigma^2 \mathbf{I}) 
\prod_{i=1}^d \exp \left\{ \tilde{v}_{1i} w_i - \frac{1}{2} \tilde{v}_{2i} w_i^2 \right\}d\mathbf{w}\,,\displaybreak[0] \label{eq:Z1} \\
\hat{Z}(\hat{v}) & = \prod_{i=1}^d \int \exp \left\{\hat{v}_{1i} w_i - \frac{1}{2}\hat{v}_{2i} w_i^2  \right\}
	f_i (w_i) d w_i\,,\displaybreak[0]\label{eq:Z2} \\
\tilde{Z}(v) &= \prod_{i=1}^d \int \exp \left\{ v_{1i} w_i - \frac{1}{2}v_{2i} w_i^2 \right\}d w_i\,,\label{eq:Z3}
\end{align}
where $\hat{\mathbf{v}}_1=(\hat{v}_{11},\ldots,\hat{v}_{1d})^\text{T}$, $\hat{\mathbf{v}}_2=(\hat{v}_{21},\ldots,\hat{v}_{2d})^\text{T}$,
$\mathbf{v}_1=(v_{11},\ldots,v_{1d})^\text{T}$, $\mathbf{v}_2=(v_{21},\ldots,v_{2d})^\text{T}$ and
the vectors $\mathbf{v}_1$ and $\mathbf{v}_2$ contain the natural parameters of the marginals of $\mathcal{Q}$.
Once a stationary point of (\ref{eq:objective}) is found, the optimal
value of $\tilde{z}_1,\ldots,\tilde{z}_d$ can be computed very easily as a
function of the solution for $\tilde{\mathbf{v}}_1$ and $\tilde{\mathbf{v}}_2$. 
See \cite{matthias2006} for further details. To find a stationary point of the objective energy, we follow \cite{IMM2005-03460}
and attempt to find a minimum of this function. However, further constraints have to be imposed on 
$\tilde{\mathbf{v}}_{2}$, $\hat{\mathbf{v}}_{2}$ and $\mathbf{v}_{2}$ to guarantee that the energy $E(v,\hat{v},\tilde{v})$ is lower bounded.

\subsection{Lower Bound on the Objective Energy}

Consider, in addition to constraints (\ref{eq:equality}), the following inequality constraints on
$\tilde{\mathbf{v}}_{2}$, $\hat{\mathbf{v}}_{2}$ and $\mathbf{v}_{2}$:
\begin{align}
\tilde{\mathbf{v}}_2 & \succeq \bm \varepsilon\,,  &  \hat{\mathbf{v}}_2 & \succeq \bm \varepsilon \,,  & \mathbf{v}_2 & \succeq 3 \bm \varepsilon \,,
\label{eq:extra_contraints}
\end{align}
where $\bm \varepsilon$ is a $d$-dimensional vector whose components are all equal to a small positive constant $\varepsilon$. 
Then, the objective (\ref{eq:objective}) is bounded from below. 
For any $\mathbf{v}_1$ and $\mathbf{v}_2$ such that $\mathbf{v}_2 \succeq 3 \bm \varepsilon$,
we can choose $\tilde{\mathbf{v}}_1 = \hat{\mathbf{v}}_1 = 0.5 \mathbf{v}_1$ and 
$\tilde{\mathbf{v}}_2 = \hat{\mathbf{v}}_2 = 0.5 \mathbf{v}_2$. These parameter values satisfy the required equality and
inequality constraints and give a lower objective than the maximizers of $E(v,\hat{v},\tilde{v})$ when $v$ is held fixed.
This is more compactly written as $\hat{v} = 0.5 v$ and $\tilde{v} = 0.5 v$, were $0.5 v = \{ 0.5 \mathbf{v}_1, 0.5 \mathbf{v}_2 \}$.
For these values of $\hat{v}$ and $\tilde{v}$ we can lower bound each term of the EP energy:
\begin{align}
- \log Z(0.5v) & \geq
	\frac{n}{2} \log (2\pi \sigma^2) - \frac{d}{2} \log (4 \pi) - \sum_{i=1}^d \left(\frac{v_{1i}^2}{4v_{2i}} - \frac{1}{2} \log v_{2i} \right)\,, \\
\log \tilde{Z}(v) & \geq  \frac{d}{2} \log(2\pi) + \sum_{i=1}^d 
	\left( \frac{v_{1i}^2}{2v_{2i}} - \frac{1}{2} \log  v_{2i}\right)\,, \\
- \log \hat{Z}(0.5v) & \geq - \frac{d}{2} \log (4\pi) - \sum_{i=1}^d \left(\frac{v_{1i}^2}{4v_{2i}} - \frac{1}{2} \log v_{2i} \right)
	- \sum_{i=1}^d \log K_i\,,
\end{align}
with $- \log K_i \geq 1/2 \log (4\pi) - 1 / 2 \log v_{2i}$. 
Since $\mathbf{v}_2 \succeq 3 \bm \epsilon$,
all these bounds are real and positive and their sum is $n/2 \log (2 \pi \sigma^2) - d / 2 \log(2)$, which is a lower bound on the
energy function minimized by the convergent EP algorithm.
Note that if the components of $\tilde{\mathbf{v}}_{2}$, $\hat{\mathbf{v}}_{2}$ and $\mathbf{v}_{2}$ could be negative,
some of the normalization constants (\ref{eq:Z1}), (\ref{eq:Z2}) and (\ref{eq:Z3}) might be infinite and the
EP energy would be unbounded. The inequality constraints in (\ref{eq:extra_contraints}) guarantee that this cannot occur.

\subsection{Provably Convergent Algorithm}

The first step of the provably convergent EP algorithm (PC-EP) 
maximizes $E(v,\hat{v},\tilde{v})$ with respect to $\hat{v}$ and $\tilde{v}$ given the current
value of $v$, which is denoted by $v^{(t)}=\{ \mathbf{v}_1^{(t)}, \mathbf{v}_2^{(t)} \}$, that is,
\begin{align}
E(v^{(t)}) &= \underset{
	\begin{array}{c}
	\hat{v},\tilde{v}\\
	\text{s.t (\ref{eq:equality}), (\ref{eq:extra_contraints})} \\
	\end{array}
	}{\text{max}} 
	\quad E(v^{(t)},\hat{v},\tilde{v})\,.
\label{eq:problem1}
\end{align}
This step is implemented using standard optimization methods such as Quasi-Newton techniques, optimizing only on either $\tilde{v}$ or $\hat{v}$ since
(\ref{eq:equality}) allows us to identify one of these tuples given the other and the current value of $v$.
The Lagrangian for this optimization problem is
\begin{align}
\mathcal{L}(v^{(t)},\hat{v},\tilde{v},\lambda,\mu) & = 
E(v^{(t)},\hat{v},\tilde{v})
+ \bm{\lambda}_1^\text{T} (\mathbf{v}_1^{(t)} - \hat{\mathbf{v}}_1 - \tilde{\mathbf{v}}_1) +
\bm{\lambda}_2^\text{T} (\mathbf{v}_2^{(t)} - \hat{\mathbf{v}}_2 - \tilde{\mathbf{v}}_2) \nonumber \\
& \quad + \bm{\mu}_1^\text{T} (\bm{\epsilon} - \hat{\mathbf{v}}_2) + \bm{\mu}_2^\text{T} (\bm{\epsilon} - \tilde{\mathbf{v}}_2)\,,
\label{eq:lagragian}
\end{align}
where $\lambda = \{ \bm \lambda_1 , \bm \lambda_2 \}$, $\mu = \{ \bm \mu_1 , \bm \mu_2 \}$ and
$\bm{\lambda}_1$, $\bm{\lambda}_2$, $\bm{\mu}_1$ and $\bm{\mu}_2$ are $d$-dimensional vectors of 
Lagrange multipliers that satisfy $\bm{\mu}_1 \preceq \mathbf{0}$ and $\bm{\mu}_2 \preceq \mathbf{0}$.
The corresponding dual function is 
\begin{align}
g(v^{(t)},\lambda,\mu) & = 
	\underset{
	\begin{array}{c}
	\hat{v},\tilde{v}
	\end{array}
	}{\text{max}} 
\mathcal{L}(v^{(t)},\hat{v},\tilde{v},\lambda,\mu)
\label{eq:dual_function}
\end{align}
The dual problem minimizes this latter function with respect to the Lagrange multipliers:
\begin{align}
g(v^{(t)}) &= 
\underset{
	\begin{array}{c}
	\lambda,\mu \\
	\text{s.t.}\, \bm{\mu}_1 \preceq \mathbf{0},\, \bm{\mu}_2 \preceq \mathbf{0}\\
	\end{array}
}{\text{min}} g(v^{(t)},\lambda,\mu)
\label{eq:dual_problem}
\end{align}
Let $\bm{\lambda}_1^\star$, $\bm{\lambda}_2^\star$, $\bm{\mu}_1^\star$ and $\bm{\mu}_2^\star$ 
be the components of $\lambda$ and $\mu$ that minimize the objective in (\ref{eq:dual_problem}) under constraints $\bm{\mu}_1 \preceq \mathbf{0}$ and
$\bm{\mu}_2 \preceq \mathbf{0}$. The second step of the convergent algorithm solves the convex problem
\begin{align}
\underset{
\begin{array}{c}
v \\
\text{s.t. $\mathbf{v}_2 \succeq 3 \bm \varepsilon$}
\end{array}
}
{\text{min}} \quad {\bm{\lambda}_1^\star}^\text{T} \mathbf{v}_1 + {\bm{\lambda}_2^\star}^\text{T} \mathbf{v}_2 + \log \tilde{Z} (v)\,.
\label{eq:problem2}
\end{align}
Let $v^{(t+1)} = \{ \mathbf{v}_1^{(t+1}), \mathbf{v}_2^{(t+1)} \}$ be the solution to this optimization problem.
When the constraint on $\mathbf{v}_2$ is not tight, we have that
$\mathbf{v}_2^{(t+1)} = (2 \bm \lambda_2^\star - \bm \lambda_1^\star \circ \bm \lambda_1^\star)^{-1}$
and $\mathbf{v}_1^{(t+1)} = -\bm \lambda_1^\star \circ \mathbf{v}_2^{(t+1)}$,
where the operator "$\circ$" denotes the Hadamard
element-wise product and the inverse of a vector is defined as a new vector whose components are the inverse of the components
of the original vector. When the constraint is tight, the solution is still given by these formulas. However,
the components of $\mathbf{v}_2$ which result to be smaller than $3\varepsilon$ (those for which the constraint is active)
must now be equal to $3\varepsilon$.

We now prove that the two steps of PC-EP, that is, (\ref{eq:problem1}) and (\ref{eq:problem2}), always generate a 
reduction in the value of $E(v^{(t)},\hat{v},\tilde{v})$.
The proof shown here is similar to the one given in \cite{IMM2005-03460} for the unconstrained case.
The objective in (\ref{eq:problem1}) is concave and the inequality constraints are affine. 
Thus, the weak form of Slater's condition is satisfied and strong duality holds.
This means that the gap between the dual and the primal problems is zero, that is, $g(v^{(t)})=E(v^{(t)})$. See \cite{Boyd2004} for further references.
Let $\lambda^\star = \{\bm{\lambda}_1^\star, \bm{\lambda}_2^\star \}$ and $\mu^\star = \{\bm{\mu}_1^\star, \bm{\mu}_2^\star \}$.
Then, we have
\begin{align}
E(v^{(t)}) =
\underset{
	\begin{array}{c}
	\hat{v}, \tilde{v}
	\end{array}
}{\text{max}} 
	\mathcal{L}(v^{(t)},\hat{v},\tilde{v}, \lambda^\star,\mu^\star)
\geq  
\underset{
	\begin{array}{c}
	\hat{v}, \tilde{v}
	\end{array}
}{\text{max}} 
	\mathcal{L}(v^{(t+1)},\hat{v},\tilde{v}, \lambda^\star,\mu^\star)\,,
\end{align}
where the first equality follows from strong duality and the following inequality
is obtained because in the second step of PC-EP we are always minimizing and consequently,
$\mathcal{L}(v^{(t)}, \hat{v}, \tilde{v}, \lambda^\star,\mu^\star)$
will never increase when we replace $v^{(t)}$ by $v^{(t+1)}$.
Continuing with the derivations, we obtain
\begin{align}
\underset{
	\begin{array}{c}
	\hat{v}, \tilde{v}
	\end{array}
}{\text{max}} 
	\mathcal{L}(v^{(t+1)},\hat{v},\tilde{v}, \lambda^\star,\mu^\star)
\geq
\underset{
\begin{array}{c}
\hat{v},\tilde{v}\\
\text{s.t (\ref{eq:equality}), (\ref{eq:extra_contraints}) } 
\end{array}
}{\text{max}}
	\mathcal{L}(v^{(t+1)},\hat{v},\tilde{v}, \lambda^\star,\mu^\star)
\geq  E(v^{(t+1)})\,,
\end{align}
where the first inequality is obtained because
adding extra constraints in a maximization problem can never produce higher values of the
target function. The second inequality is obtained by expanding $\mathcal{L}(v^{(t+1)},\hat{v},\tilde{v}, \lambda^\star,\mu^\star)$
according to (\ref{eq:lagragian}), using the fact that $\bm{\mu}_1^T (\bm{\epsilon}-\tilde{\mathbf{v}}_2)$ 
and $\bm{\mu}_2^T (\bm{\epsilon}-\hat{\mathbf{v}}_2)$ can only be positive and finally, using the definition (\ref{eq:problem1}).

Note that we still need the vectors $\bm{\lambda}_1^\star$ and $\bm{\lambda}_2^\star$ 
for the practical implementation of the second step of PC-EP. 
Let $\hat{v}^\star = \{\hat{\mathbf{v}}_1^\star, \hat{\mathbf{v}}_2^\star \}$ and
$\tilde{v}^\star = \{\tilde{\mathbf{v}}_1^\star, \tilde{\mathbf{v}}_2^\star \}$ be the 
solution to (\ref{eq:problem1}). Then, the gradient of $\mathcal{L}$ with respect to 
the elements in $\tilde{v}$ and $\hat{v}$ should be zero at
$\tilde{v} = \tilde{v}^\star$, $\hat{v} = \hat{v}^\star$,
$\lambda = \lambda^\star$ and $\mu = \mu^\star$.
This generates the equations
\begin{align}
\lambda_{1i}^\star & = -\mathds{E}_\mathcal{Q}[w_i] = -\mathds{E}_\mathcal{P}[w_i]\,, &
\lambda_{2i}^\star + \mu_{1i}^\star & = \frac{1}{2}\mathds{E}_\mathcal{P}[w_i^2] \,, &
\lambda_{2i}^\star + \mu_{2i}^\star & = \frac{1}{2}\mathds{E}_\mathcal{Q}[w_i^2] \,,\label{eq:valueLambdaStar}
\end{align}
for $i=1,\ldots,d$, where $\mathds{E}_\mathcal{Q}$ denotes expectation with respect to the normalized
version of $\mathcal{Q}$ and $\mathds{E}_\mathcal{P}$ denotes expectation with respect to 
$\mathcal{P}(w_i) \propto \exp \{ \hat{v}_{1i} w_i - \frac{1}{2} \hat{v}_{2i} w_i^2 \}f_i(w_i)$.
The inequalities $\tilde{\mathbf{v}}_2 \succeq \bm \varepsilon$ and $\hat{\mathbf{v}}_2 \succeq \bm \varepsilon$
can only be active on either $\hat{v}_{2i}^\star$ or $\tilde{v}_{2i}^\star$, but not on both
at the same time. The reason for this is that $\hat{\mathbf{v}}_2$ and $\tilde{\mathbf{v}}_2$ must satisfy
$\hat{\mathbf{v}}_2 + \tilde{\mathbf{v}}_2 = \mathbf{v}_2$ and $\mathbf{v}_2$ satisfies $\mathbf{v}_2 \succeq 3 \bm \varepsilon$.
This means that, for each $i$, only either $\mu_{1i}^\star$ or $\mu_{2i}^\star$ can be different from zero, but not both at the same time.
Therefore, when $\tilde{v}_{2i}^\star = \varepsilon$, we have that $\mu_{1i}^\star = 0$ and
$\lambda_{2i}^\star = 1 / 2 \mathds{E}_\mathcal{P}[w_i^2]$. On the other hand, when $\hat{v}_{2i}^\star = \varepsilon$, we have that $\mu_{2i}^\star = 0$ and
$\lambda_{2i}^\star = 1 / 2 \mathds{E}_\mathcal{Q}[w_i^2]$.
Finally, when no inequality constraints are active in $\hat{\mathbf{v}}_2^\star$ and $\tilde{\mathbf{v}}_2^\star$, that is,
$\bm \mu_1^\star = \bm 0$ and $\bm \mu_2^\star = \bm 0$, we have that 
$\lambda_{2i}^\star = 1 / 2 \mathds{E}_\mathcal{Q}[w_i^2] = 1 / 2 \mathds{E}_\mathcal{P}[w_i^2]$ for $i = 1,\ldots,d$.
The expectations with respect to $\mathcal{Q}$ are given by
\begin{align}
\mathds{E}_\mathcal{Q}[w_i] & = m_{i}^\star \,, & \mathds{E}_\mathcal{Q}[w_i^2] & = (\mathbf{A}^\star)^{-1}_{ii}\,,
\end{align}
for $i = 1,\ldots,d$, where $\mathbf{A}^\star = \sigma^{-2} \mathbf{X}^\text{T} \mathbf{X} + \text{diag}(\tilde{\mathbf{v}}_2^\star)$ and
$\mathbf{m}^\star = (\mathbf{A}^\star)^{-1} (\tilde{\mathbf{v}}_1^\star + \sigma^{-2} \mathbf{X}^\text{T}\mathbf{y})$.
For computing the expectations with respect to $\mathcal{P}$, we define 
\begin{align}
p_i & = -\frac{1}{2}\log(\hat{v}_{2i}^\star) - \frac{1}{2}\log((\hat{v}_{2i}^\star)^{-1} + v_s) +
\frac{1}{2} \frac{ (\hat{v}_{1i}^\star)^2}{(\hat{v}_{2i}^\star)^2} \left[\hat{v}_{2i}^\star - ((\hat{v}_{2i}^\star)^{-1}+v_s)^{-1}\right]\,,\\
a_i & = \sigma(p_i + \rho_s) \frac{\hat{v}^\star_{1i}}{1 + \hat{v}^\star_{2i}v_s} +
\sigma(-p_i- \rho_s) \hat{v}^\star_{1i}\,,\\
b_i & = \sigma(p_i + \rho_s) \frac{ (\hat{v}^\star_{1i})^2 (\hat{v}^\star_{2i})^{-2} - (\hat{v}^\star_{2i})^{-1} - v_s}{((\hat{v}^\star_{2i})^{-1} + v_s)^{2}} +
\sigma(-p_i - \rho_s) \left[ (\hat{v}^\star_{1i})^2 - \hat{v}^\star_{2i} \right]\label{eq:bi:chap:4}\,.
\end{align}
where $\rho_s = \log(p_s) - \log(1 - p_s)$ and $\sigma$ denotes the logistic function.
The expectations with respect to $\mathcal{P}$ are then given by
\begin{align}
\mathds{E}_\mathcal{P}[w_i] & = (\hat{v}_{2i}^\star)^{-1} (\hat{v}_{1i}^\star - a_i) \,, &
\mathds{E}_\mathcal{P}[w_i^2] & = (\hat{v}_{2i}^\star)^{-1} - (\hat{v}_{2i}^\star)^{-2}(a_i^2 - b_i) + \mathds{E}_\mathcal{P}[w_i]^2 \,,
\end{align}
for $i = 1,\ldots,d$. See \cite{jmhlPhd2010} for the derivation of these formulas.

The cost of PC-EP is determined by the computation of the diagonal of the covariance matrix of
the posterior approximation, that is, matrix $\mathbf{A}^{-1}$ in (\ref{eq:Q}). 
When the number $d$ of features is larger than the number $n$ of training instances (that is, $d \gg n$), we can use the Woodbury formula
to compute $\mathbf{A}^{-1}$ in $\mathcal{O}(dn^2)$ operations. This is the same computational cost of regular EP \cite{jmhlPhd2010} since this method
also has to compute the diagonal of this matrix on each iteration. However, PC-EP computes the marginal variances of the posterior approximation
each time that the gradient of the objective in (\ref{eq:problem1}) needs to be evaluated.
This means that the multiplicative constant hidden in the cost of PC-EP is about two orders of magnitude larger than the one hidden in the cost of regular EP.
Nevertheless, by applying the optimization algorithm described in \cite{Seeger2011} to PC-EP, this method can actually become as fast as regular EP.
This is left as future work.

\section{Experiments}

The performance of regular EP (R-EP) with damping to improve convergence is compared with the proposed 
provably convergent algorithm (PC-EP). For this, we consider a synthetic 
linear regression problem with $d=25$ dimensions and $\mathbf{w}$ sampled according to (\ref{eq:spike_and_slab}). 
Only $5$ components of $\mathbf{w}$ are different from zero on average ($p_s = 0.2$). These are drawn from a standard Gaussian 
distribution ($v_s = 1$). The attribute vectors $\mathbf{x}_i$ are uniformly sampled from the 
unit hyper-sphere. The targets are sampled according to
$y_i = \mathbf{w}^T \mathbf{x}_i + \epsilon_i$, where $\epsilon_i \sim \mathcal{N}(0,\sigma^2)$ and
$\sigma = 0.005$. Using these settings, we generate 100 training and 
test sets with 10 and 1000 instances, respectively.

R-EP and PC-EP are run on each training set and their prediction performance
is evaluated on the corresponding test set. The method R-EP is run for a maximum of $1000$ iterations using
the same constraints (\ref{eq:extra_contraints}) enforced in PC-EP. 
Different values are considered
for the damping parameter: $\tau \in \{0.1, 0.3, 0.5, 0.7, 0.9\}$.
The table in the left-hand side of Figure \ref{tab:MSE_not_converges} shows, for each value of $\tau$, the average mean square error (MSE) of each 
method on the test sets for which R-EP does not converge on the associated training sets. 
The number of these test sets is also displayed in the table. The other table in the right-hand side of the figure shows the same information
for those sets in which R-EP does converge.
These results indicate that, when R-EP does not converge, the prediction errors of this method are typically larger than when it does
converge. However, more importantly, when R-EP does not converge, the posterior approximation generated by PC-EP is better than 
the one obtained by R-EP. On the other hand, when R-EP does converge,
the prediction errors of both methods are significantly lower and both techniques obtain similar performances.

\begin{figure}[bt!]
\caption{Left, average mean square error (MSE) of R-EP and PC-EP on the test sets for which R-EP does {\bf not} converge on the associated training sets.
Right, average MSE of R-EP and PC-EP on the test sets for which R-EP does converge.}
\vspace{0.5cm}
\begin{minipage}{0.5\linewidth}
\label{tab:MSE_not_converges}
\begin{center}
\begin{tabular}{r@{$\pm$}lr@{$\pm$}lcc}
\hline
\multicolumn{6}{c}{\bf{R-EP does not Converge}} \\
\hline
\multicolumn{2}{c}{\bf{PC-EP}} & 
\multicolumn{2}{c}{\bf{R-EP}} & 
$\tau$ & \bf{\# Sets}\\
\hline
0.24 & 0.09 & 0.29 & 0.09 &  0.1 &    6 \\
0.16 & 0.04 & 0.20 & 0.05 &  0.3 &   10 \\
0.14 & 0.03 & 0.27 & 0.05 &  0.5 &   13 \\
0.18 & 0.05 & 0.22 & 0.06 &  0.7 &   14 \\
0.15 & 0.04 & 0.17 & 0.05 &  0.9 &   17 \\
\hline
\end{tabular}
\end{center}
\end{minipage}
\begin{minipage}{0.5\linewidth}
\begin{center}
\begin{tabular}{r@{$\pm$}lr@{$\pm$}lcc}
\hline
\multicolumn{6}{c}{\bf{R-EP does Converge}} \\
\hline
\multicolumn{2}{c}{\bf{PC-EP}} & 
\multicolumn{2}{c}{\bf{R-EP}} & 
$\tau$ & \bf{\# Sets}\\
\hline
0.041 & 0.007 & 0.040 & 0.009 &   0.1 &   94 \\
0.041 & 0.009 & 0.045 & 0.012 &   0.3 &   90 \\
0.040 & 0.010 & 0.042 & 0.012 &   0.5 &   87 \\
0.032 & 0.007 & 0.029 & 0.008 &   0.7 &   86 \\
0.033 & 0.007 & 0.024 & 0.008 &   0.9 &   83 \\
\hline
\end{tabular}
\end{center}
\end{minipage}
\end{figure}

\section{Conclusions}

The LRMSSP is interesting in under-determined scenarios with $n \ll d$ since the spike-and-slab prior can help to alleviate over-fitting in these cases.
Nevertheless, approximate inference in this model can be difficult. More precisely, expectation propagation 
(EP) may have convergence problems in some extreme cases and one is forced to limit the maximum number 
of iterations of the algorithm. In this work, we have proposed a provably convergent EP algorithm (PC-EP) for the LRMSSP. Each iteration of PC-EP
is proved to minimize an energy function whose stationary points coincide with the solution of regular EP (R-EP). 
We also introduce constraints on the parameters of the EP approximation
so that this energy function is bounded, which guarantees convergence. Experiments with synthetic
data illustrate the advantages of PC-EP over R-EP. When R-EP does not converge,
the posterior approximation given by PC-EP is usually better.
By contrast, when R-EP does converge, both methods perform similarly.

\bibliography{Bibliography}

\bibliographystyle{unsrt}

\end{document}